\documentclass[11pt,twocolumn]{article}

\usepackage[utf8]{inputenc}
\usepackage[T1]{fontenc}
\usepackage{times}
\usepackage{microtype}
\usepackage{amsmath,amssymb}
\usepackage{graphicx}
\usepackage{booktabs}
\usepackage{listings}
\usepackage{xcolor}
\usepackage{hyperref}
\usepackage{cleveref}
\usepackage{url}
\usepackage{enumitem}
\usepackage[utf8]{inputenc}
\usepackage[T1]{fontenc}
\usepackage{amsmath}
\usepackage{graphicx}
\usepackage[margin=1in]{geometry}
\hypersetup{
  colorlinks=true,
  linkcolor=blue,
  citecolor=blue,
  urlcolor=blue,
  pdfauthor={Chao Li},
  pdftitle={Domain-Contextualized Concept Graphs}
}

\title{\textbf{Domain-Contextualized Concept Graphs: \\
A Computable Framework for Knowledge Representation}}

\author{
  \textbf{Chao Li} \\
  Deepleap.ai \\
  \texttt{lichao@deepleap.ai }
  \and
  \textbf{Yuru Wang} \\
  School of Information Science and Technology, Northeast Normal University \\
  \texttt{wangyr915@nenu.edu.cn}
}

\date{}

\begin{document}

\maketitle

\begin{abstract}
Traditional knowledge graphs are constrained by fixed ontologies that organize concepts within rigid hierarchical structures. The root cause lies in treating \textbf{domains as implicit context} rather than as explicit, reasoning-level components. To overcome these limitations, we propose the \textbf{Domain-Contextualized Concept Graph (CDC)}, a novel knowledge modeling framework that elevates domains to first-class elements of conceptual representation. CDC adopts a \textbf{C–D–C triple structure}, $\langle$\textit{Concept}, \textit{Relation}@\textit{Domain}, \textit{Concept'}$\rangle$, where domain specifications serve as dynamic classification dimensions defined on demand. Grounded in a \textbf{cognitive–linguistic isomorphic mapping principle}, CDC operationalizes how humans understand concepts through contextual frames. We formalize 20+ standardized relation predicates (structural, logical, cross-domain, and temporal) and implement CDC in Prolog for full inference capability. Case studies in education, enterprise knowledge systems, and technical documentation demonstrate that CDC enables context-aware reasoning, cross-domain analogy, and personalized knowledge modeling—capabilities unattainable under traditional ontology-based frameworks.
\end{abstract}


\section{Introduction}

\subsection{The Fundamental Limitation of Fixed Ontologies}

Knowledge graphs have revolutionized how we organize and query information, powering systems from Google’s search to biomedical research databases~\cite{hogan2021knowledge}. Yet they suffer from a fundamental architectural constraint: \textbf{concepts must be classified into fixed ontological categories that apply universally}. This rigidity simplifies reasoning within a single ontology but severely limits expressiveness when knowledge must adapt to different domains, learners, or historical contexts.

\paragraph{Scenario 1: Cross-Disciplinary Research.} 
A neuroscientist studying artificial intelligence may wish to ask: “What structural similarities exist between artificial neural networks and biological neural systems?”  

In traditional knowledge graphs~\cite{auer2007dbpedia,bollacker2008freebase}, this query requires manual ontology alignment across disciplines. The computer science ontology defines a neural network as an algorithm, while the neuroscience ontology treats the brain as an organ. Even when bridge ontologies~\cite{euzenat2013ontology} are introduced, the expressive power remains limited—relations such as “analogous to’’ cannot be formally represented, only identity or subsumption. As a result, the two domains cannot communicate within the graph itself; cross-domain insights remain confined to human interpretation rather than machine reasoning.

\paragraph{Scenario 2: Personalized Learning.} 
An educational platform teaching “machine learning’’ to students from diverse backgrounds faces the same rigidity. 

The concept must shift in meaning—function approximation for mathematicians, brain-like learning for biologists, and automated pattern recognition for engineers. Traditional knowledge graphs cannot support such contextual variation. They either impose one canonical definition, losing adaptability, or maintain disjoint ontologies, fragmenting the conceptual space and preventing reasoning across perspectives.

\paragraph{Scenario 3: Temporal Knowledge Evolution.} 
The concept of the “atom’’ has evolved dramatically~\cite{flouris2008ontology}. Dalton’s model in the 1800s viewed it as an indivisible particle; Thomson’s early twentieth-century theory introduced internal structure; Bohr’s model added quantized energy levels; and modern physics conceives it as a quantum field excitation. Conventional ontology-based approaches cannot coherently capture this evolution: they either overwrite past definitions or create disconnected entities such as \texttt{Atom\_Dalton} and \texttt{Atom\_Bohr}, thereby losing the semantic continuity that links successive theoretical stages.  

\subsection{The Root Cause}

These limitations stem from treating domains as \textit{implicit context} rather than \textit{explicit structural elements}. When DBpedia encodes \texttt{(Apple, instance-of, Company)}, the domain context “technology industry’’ exists only in human interpretation, not in the formal structure. The same graph cannot determine when “Apple’’ refers instead to a fruit. Because domains remain unmodeled, knowledge graphs cannot represent when, where, or for whom a given relation applies. Consequently, systems that require cross-domain analogy, contextual adaptation, or temporal reasoning must rely on costly ontology engineering and manual schema alignment. Fixed ontologies thus fail to meet the needs of modern AI systems that demand dynamic, context-aware, and interdisciplinary reasoning.

\subsection{Theoretical Foundation}

CDC is grounded in a principle from cognitive science: \textbf{humans understand concepts through domain-dependent frames}~\cite{fillmore1982frame,barsalou1982context}. The meaning of “bank’’ in the domains of finance versus geography is not mere lexical ambiguity; it reflects how cognition contextualizes concepts~\cite{gardenfors2000conceptual}. This insight, combined with formal approaches to conceptual structure, motivates CDC’s design.  

We formalize this through a \textbf{three-level isomorphic mapping}:

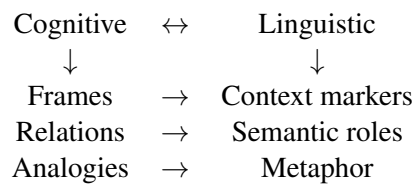
\begin{center}
\begin{tabular}{ccc}
Cognitive & $\leftrightarrow$ & Linguistic \\
$\downarrow$ & & $\downarrow$ \\
Frames & $\rightarrow$ & Context markers \\
Relations & $\rightarrow$ & Semantic roles \\
Analogies & $\rightarrow$ & Metaphor
\end{tabular}
\end{center}

This isomorphism ensures that CDC's computational structure faithfully represents cognitive organization~\cite{lakoff1980metaphors}. The C-D-C triple $\langle$Concept, Relation@Domain, Concept'$\rangle$ mirrors how humans naturally encode contextualized conceptual relationships.

\subsection{Key Design Principles}

CDC embodies four interrelated principles that collectively ensure flexibility, consistency, and computability.  

\paragraph{Domain Flexibility.}  
Domains are defined on demand as dynamic classification dimensions rather than fixed taxonomic categories. A specification such as \texttt{'HighSchool@Math@Calculus'} simply scopes the applicability of a relation without constraining its semantics.  

\paragraph{Relation Standardization.}  
Although domains are flexible, relations are standardized. The CDC framework defines more than twenty core relation predicates (e.g., \texttt{is\_a}, \texttt{part\_of}, \texttt{requires}) with precise formal semantics and computational properties.  

\paragraph{Full Computability.}  
Every CDC relation can be expressed as a Prolog predicate~\cite{kowalski1974predicate} with explicit inference rules. This enables automated reasoning, consistency checking, and domain-sensitive query answering.  

\paragraph{Cross-Domain Reasoning.}  
CDC supports relations that connect concepts across domains, including \texttt{analogous\_to@D$_1$\,$\leftrightarrow$\,D$_2$} (for structural similarity) and \texttt{fuses\_with@D$_1$\,$\oplus$\,D$_2$} (for conceptual integration). These operators allow knowledge graphs to represent analogies and hybridization processes within a single coherent framework.

\subsection{Contributions}

This paper makes four primary contributions. 
\begin{itemize}
    \item It introduces the \textbf{Domain-Contextualized Concept Graph (CDC)} framework, a cognitive-linguistically grounded model that treats domains as explicit structural components rather than implicit contextual assumptions. 
    \item It develops a \textbf{standardized predicate system} comprising more than twenty formally defined relations across structural, logical, cross-domain, temporal, and degree dimensions. 
    \item It provides a \textbf{computational implementation} of CDC in Prolog, including the core predicates, inference mechanisms, and query interfaces necessary for automated reasoning. 
    \item It validates the framework through \textbf{methodological case studies} in education, enterprise knowledge management, and technical documentation, demonstrating CDC’s unified capacity for adaptive and context-aware knowledge modeling.
\end{itemize}

\section{Related Work}

CDC builds upon, integrates, and diverges from several major research traditions, including knowledge graph systems, ontology engineering, cognitive science, and logic-based representation. This section situates CDC in relation to these foundational strands.

\subsection{Knowledge Graph Systems}

Early large-scale knowledge graphs such as DBpedia~\cite{auer2007dbpedia} and Freebase~\cite{bollacker2008freebase} established the paradigm of representing world knowledge through fixed ontological schemas. While effective for large-scale information retrieval, these systems are inherently limited in expressing context-dependent meanings; for example, the triple \texttt{(Neural\_Network, instance\_of, Algorithm)} applies globally and cannot be scoped to particular interpretive contexts.

Wikidata~\cite{vrandevcic2014wikidata} introduced partial flexibility through the use of \textit{qualifiers}, allowing statements to include temporal or situational metadata. However, these qualifiers are annotations \textit{about} statements rather than structural components of reasoning. In contrast, CDC treats domains as first-class semantic entities that define the scope of relations and enable cross-domain inference.  
Similarly, Schema.org~\cite{guha2016schema} provides a widely used structured vocabulary for web markup but remains web-centric and primarily optimized for search-engine interoperability rather than general-purpose knowledge representation.

\subsection{Ontology Engineering}

Efforts in ontology engineering have produced a spectrum of approaches from highly abstract upper ontologies to specialized domain ontologies. Upper ontologies such as SUMO~\cite{niles2001towards}, Cyc~\cite{lenat1995cyc}, and BFO~\cite{arp2015building} seek comprehensive axiomatic foundations that assume universal categories applicable across all knowledge domains. These systems provide formal rigor and ontological consistency~\cite{guarino2009ontology} but often sacrifice adaptability. CDC takes an alternative stance, favoring flexibility and contextual alignment over universality.

At the other end of the spectrum, domain-specific ontologies such as SNOMED CT and the Gene Ontology~\cite{smith2007obo} deliver rich internal structure within their respective fields but are notoriously difficult to integrate across disciplinary boundaries. CDC positions itself as a cross-domain layer that connects such specialized ontologies through explicit domain contextualization.

\subsection{Cognitive Science Foundations}

The design of CDC is informed by theoretical work in cognitive science and linguistics. Gärdenfors’ theory of conceptual spaces~\cite{gardenfors2000conceptual} models concepts as regions within multidimensional quality spaces, offering an account of graded similarity and prototype effects.  
Fillmore’s frame semantics~\cite{fillmore1982frame} further suggests that meanings are interpreted relative to conceptual frames, a principle that CDC operationalizes computationally by embedding domain-specific frames into the graph structure.  
Related studies on context-dependent categorization~\cite{barsalou1982context,medin1988context} demonstrate that human cognition systematically adapts categorization to situational context. CDC extends this insight by formalizing such adaptive semantics within a computational reasoning framework.

\subsection{Logic-Based Representation}

CDC also relates to formal logic traditions in knowledge representation. Description Logic (DL) provides the semantic foundation of OWL ontologies~\cite{baader2003description,horrocks2003shiq}, emphasizing decidable reasoning under well-defined constraints. In contrast, CDC employs Prolog as its computational substrate, prioritizing expressive power and contextual inference over guaranteed decidability.

Another relevant line of work is RDF named graphs~\cite{carroll2005named}, which extend triples to quads for tracking provenance and statement grouping. However, named graphs primarily serve as technical containers for metadata rather than semantic contexts that influence reasoning. CDC diverges fundamentally by treating domains as semantic contexts that directly participate in logical inference.

\subsection{Summary}

Table~\ref{tab:related} summarizes how CDC relates to these major approaches across different representational layers. In essence, CDC complements traditional knowledge graph systems by introducing an explicit notion of domain context that bridges formal ontology, cognitive grounding, and computational reasoning.

\begin{table*}[t]
\centering
\scriptsize
\begin{tabular}{p{2.5cm}p{2cm}p{3.5cm}p{4cm}}
\toprule
\textbf{Approach} & \textbf{Layer} & \textbf{Goal} & \textbf{CDC's Relationship} \\
\midrule
DBpedia/Freebase & Implementation & Web-scale extraction & Provides a data substrate \\
Wikidata & Implementation & Collaborative curation & Extends with domain-scoped reasoning \\
Upper ontologies & Formal semantics & Axiomatic foundations & Adopts a more flexible philosophy \\
Domain ontologies & Specialized & Deep domain coverage & Acts as a cross-domain integration layer \\
Conceptual spaces & Geometric theory & Similarity and prototypes & Provides cognitive inspiration \\
Frame semantics & Linguistic theory & Context-dependent meaning & Offers theoretical grounding \\
Description logic & Logical formalism & Decidable reasoning & Trades decidability for expressiveness \\
Named graphs & Technical & Statement grouping and provenance & Reinterprets as semantic context \\
\bottomrule
\end{tabular}
\caption{CDC's positioning relative to major approaches.}
\label{tab:related}
\end{table*}

\section{The CDC Modeling Approach}

CDC reconceptualizes knowledge representation by treating domains as first-class structural elements. 
This section begins with a motivating example that illustrates the limitations of existing approaches, followed by the theoretical principle guiding CDC’s design, the formal definition of its structure, domain specification patterns, and its expressive power.

\subsection{Motivating Example}

To motivate the design of CDC, consider constructing a knowledge system for an interdisciplinary AI course taught to students from diverse backgrounds. 
The instructor needs to explain ``neural networks'' differently depending on each student's context. 
For biology students, neural networks are computational models \textit{analogous to} biological neural systems; for mathematics students, they are parameterized function approximators; for philosophy students, they raise questions about embodied cognition and symbol grounding.

Traditional knowledge graphs would require three separate ontologies (biology, mathematics, and philosophy), manual creation of alignment rules between ontologies, and complex reasoning to determine which perspective applies to which student. 
They also cannot easily express cross-domain relationships such as ``analogous to'' across ontological boundaries. 
This approach becomes untenable when scaling to thousands of concepts and dozens of perspectives. 
CDC addresses this challenge by making domains explicit structural elements within the representation itself.

\subsection{Core Principle}

\subsubsection{Isomorphic Mapping Framework}

CDC's design is motivated by a principle from cognitive science and linguistics: \textbf{human conceptual understanding exhibits structural correspondence across three levels}---cognitive organization, linguistic expression, and formal computation~\cite{goldberg2006constructions}. 
This isomorphic mapping ensures that computational structures in CDC mirror the way humans interpret meaning across contextual frames.

\subsection{The C–D–C Structure}

\subsubsection{Formal Definition}

\textbf{Definition 1 (CDC Triple):} A CDC triple is a four-tuple:
\[
\tau = \langle c, r, c', d \rangle
\]
where:
\begin{itemize}[leftmargin=*]
\item $c \in C$ (source concept)
\item $r \in R$ (relation predicate)
\item $c' \in C$ (target concept)
\item $d \in D$ (domain specification)
\end{itemize}

\textbf{Notation:} $c \xrightarrow{r@d} c'$ or $r(c, c', d)$

\textbf{Key Distinction:} Traditional KG: $\langle$subject, predicate, object$\rangle$ (domain implicit).  
CDC: $\langle$concept, relation, concept', domain$\rangle$ (domain explicit).

\subsubsection{Domain Space}

\textbf{Definition 2 (Domain Specification):} A domain specification $d \in D$ is a structured string:
\[
d := \text{dimension} \mid \text{dimension}@d
\]

\textbf{Examples:}
\begin{itemize}[leftmargin=*]
\item Single: \texttt{'Physics'}
\item Two: \texttt{'Physics@Quantum\_Mechanics'}
\item Three: \texttt{'HighSchool@Math@Calculus'}
\item Complex: \texttt{'Student\_Zhang@Grade10'}
\end{itemize}

\textbf{Critical Property:} Domains are \textbf{defined on-demand}, not from a fixed taxonomy.

\subsection{Domain Patterns}

Table~\ref{tab:domain-patterns} shows application-specific patterns used in different contexts.

\begin{table*}[t]
\centering
\scriptsize
\begin{tabular}{p{3cm}p{4cm}p{5cm}}
\toprule
\textbf{Use Case} & \textbf{Pattern} & \textbf{Example} \\
\midrule
Academic Research & \texttt{Discipline@Theory} & \texttt{'Psychology@Behaviorism'} \\
Education & \texttt{Grade@Subject@Topic} & \texttt{'HighSchool@Chemistry@Organic'} \\
Enterprise & \texttt{Department@Project} & \texttt{'Engineering@ProductA@Testing'} \\
Historical & \texttt{Era@Region@Movement} & \texttt{'Renaissance@Italy@Humanism'} \\
Technical Docs & \texttt{Stack@Version@Context} & \texttt{'React@18.x@Mobile\_Apps'} \\
Personal Learning & \texttt{Individual@Background} & \texttt{'Student\_Li@CS\_Major'} \\
\bottomrule
\end{tabular}
\caption{Domain definition patterns.}
\label{tab:domain-patterns}
\end{table*}

\subsubsection{Design Principles}

\textbf{Principle 1: Specificity} — Domains should be fine-grained enough to disambiguate concepts but not over-fragment knowledge.

\textbf{Principle 2: Consistency} — Domain patterns should remain internally consistent across the knowledge base.

\textbf{Principle 3: Compositionality} — Hierarchical domain composition should be used whenever natural and meaningful.

\textbf{Principle 4: Context-Sensitivity} — Domains should capture the most relevant distinguishing context rather than redundant metadata.

\subsection{CDC Representation in Practice}

CDC extends the traditional triple $\langle$subject, predicate, object$\rangle$ into a four-tuple $\langle$concept, relation, concept', domain$\rangle$, where domains explicitly scope the applicability of relations. 
This allows multiple, potentially divergent categorizations of the same concept to coexist without contradiction.

For instance, the following representation distinguishes between the biological and business interpretations of ``Apple'':
\begin{lstlisting}[language=Prolog]
is_a(Apple, Fruit, 
     'Biology@Plant_Taxonomy').
is_a(Apple, Company, 
     'Business@Technology_Industry').
\end{lstlisting}

Similarly, the interdisciplinary ``neural network'' concept can be contextualized across domains:
\begin{lstlisting}[language=Prolog]
% Same concept, different domains
is_a(Neural_Network, 
     Computational_Model, 
     'CS@ML').
analogous_to(Neural_Network, 
     Biological_Brain, 
     'CS@ML', 
     'Biology@Neuroscience').
is_a(Neural_Network, 
     Function_Approximator, 
     'Math@Optimization').
is_a(Neural_Network, 
     Philosophy_Topic, 
     'Philosophy@Mind').

% Student-specific context
context_value(Student_Profile, 
     Biology_Background, 
     'Student_Zhang').
strategy(Teach_Neural_Networks, 
     'Start_with_biological_analogy', 
     'Student_Zhang@Biology').
\end{lstlisting}

Because domains are treated as first-class entities, the knowledge system can reason contextually—querying within a specific domain, linking analogical relations across domains, and generating personalized explanations for individual learners—while preserving a unified conceptual structure.

\subsection{Expressive Power}

\subsubsection{Fundamental Theorem}

\textbf{Theorem 1 (Domain Separation):} In CDC, a concept $c$ can have distinct, non-contradictory categorizations:
\[
\forall c \in C, \forall d_1, d_2 \in D, \forall r \in R, \forall c'_1, c'_2 \in C:
\]
\[
[r(c, c'_1, d_1) \land r(c, c'_2, d_2) \land c'_1 \neq c'_2 \land d_1 \neq d_2] \text{ is consistent.}
\]

\textbf{Example:}
\begin{lstlisting}[language=Prolog]
is_a(Apple, Fruit, 
     'Biology@Plant_Taxonomy').
is_a(Apple, Company, 
     'Business@Tech_Sector').
\end{lstlisting}

These coexist because domains provide separate semantic namespaces, allowing knowledge evolution and contextual adaptation without contradiction.

CDC unifies conceptual diversity under a domain-explicit formalism that preserves contextual distinctions without fragmenting the knowledge base. 
Its representational scheme integrates cognitive insight, formal logic, and computational implementability, forming the foundation upon which the predicate system and reference implementation described in the following sections are built.

\section{Relation Predicate System}

CDC’s expressive power arises not only from its domain-explicit representation but also from the design of a concise and semantically well-structured relation predicate system. This section presents the guiding philosophy, core relation categories, formal properties, and representative inference rules implemented in Prolog.

\subsection{Design Philosophy}

The CDC relation system is designed according to four key principles that balance expressiveness, computational tractability, and semantic clarity.

\paragraph{Principle 1: Compact Core.}  
CDC adopts a compact relational vocabulary of approximately twenty predicates, in contrast to Wikidata’s more than 9,000 properties. The goal is to achieve conceptual parsimony while retaining sufficient expressiveness for domain reasoning.

\paragraph{Principle 2: Semantic Orthogonality.}  
Each relation encodes a distinct semantic function. Redundant or overlapping relations are intentionally avoided to ensure interpretability and clean inferential behavior.

\paragraph{Principle 3: Formal Specification.}  
Every predicate is formally defined in terms of its arity, algebraic properties (e.g., symmetry, transitivity, reflexivity), and computational complexity, allowing predictable reasoning semantics.

\paragraph{Principle 4: Cross-Domain Support.}  
Unlike traditional systems restricted to intra-domain reasoning, CDC includes first-class predicates that operate across domains, explicitly linking heterogeneous conceptual spaces.

\subsection{Core Relations}

CDC organizes its core relations into three major categories: structural, logical, and cross-domain.

\subsubsection{Structural Relations}

Structural relations describe taxonomic and mereological organization within a domain.

\paragraph{\texttt{is\_a@D} (Taxonomic Classification)}  
\textbf{Signature:} \texttt{is\_a(C, C', D)}  
\textbf{Properties:} Transitive, Asymmetric

\textbf{Inference Rules:}
\begin{lstlisting}[language=Prolog]
% Transitive closure
is_a_transitive(X, Z, D) :- 
    is_a(X, Y, D), 
    is_a(Y, Z, D).

% Attribute inheritance
has_attribute(X, Attr, D) :-
    is_a(X, Y, D),
    has_attribute(Y, Attr, D).
\end{lstlisting}

\paragraph{\texttt{part\_of@D} (Mereological Relation)}  
\textbf{Properties:} Transitive, Asymmetric.  
Represents compositional relationships such as subsystem inclusion or part–whole hierarchies.

\paragraph{\texttt{has\_attribute@D} (Property Relation)}  
\textbf{Properties:} Neither transitive nor symmetric.  
Used to associate entities with descriptive features without implying hierarchical structure.

\subsubsection{Logical Relations}

Logical relations encode dependency, causality, and contrast relations among concepts.

\paragraph{\texttt{requires@D} (Prerequisite Relation)}  
\textbf{Properties:} Transitive, Asymmetric, Acyclic.  
Indicates that one concept must be understood or achieved before another.

\textbf{Inference Rules:}
\begin{lstlisting}[language=Prolog]
% Prerequisite chain
prerequisite_chain(X, Z, D) :-
    requires(X, Y, D),
    prerequisite_chain(Y, Z, D).
prerequisite_chain(X, Y, D) :-
    requires(X, Y, D).
\end{lstlisting}

\paragraph{\texttt{cause\_of@D} and \texttt{enables@D}}  
Represent causal and facilitative dependencies.  
While \texttt{cause\_of@D} denotes direct causal influence, \texttt{enables@D} captures enabling conditions without strict causality.

\paragraph{\texttt{contrasts\_with@D} (Oppositional Relation)}  
\textbf{Properties:} Symmetric, Non-transitive.  
Encodes conceptual oppositions such as “supervised” vs. “unsupervised” learning within the same domain.

\subsubsection{Cross-Domain Relations}

CDC introduces two unique predicates that support reasoning across domain boundaries, enabling analogical and integrative knowledge construction.

\paragraph{\texttt{analogous\_to@D\_1<->D\_2} (Structural Analogy)}
\textbf{Signature:} analogous\_to(C\_1, C\_2, D\_1, D\_2)
\textbf{Properties:} Symmetric.  
This predicate captures structural similarity between concepts that belong to distinct domains.

\textbf{Examples:}
\begin{lstlisting}[language=Prolog]
analogous_to(Atom, Solar_System, 
    'Physics@Atomic', 
    'Astronomy@Planetary').

analogous_to(Neural_Network, 
    Brain, 
    'CS@ML', 
    'Neuroscience@Cognition').
\end{lstlisting}

\paragraph{\texttt{fuses\_with@D\_1+ D\_2} (Conceptual Integration)} 
\textbf{Signature:} fuses\_with(C\_1, C\_2, NewC, D)
Represents the fusion of concepts from different domains to form a new integrative entity.

\textbf{Example:}
\begin{lstlisting}[language=Prolog]
fuses_with(User_Experience, 
    Technical_Feasibility, 
    Product_Design, 
    'UX+Engineering').
\end{lstlisting}

\subsection{Properties Summary}

Table~\ref{tab:relation-properties} summarizes the formal properties of representative CDC relations. The table distinguishes transitivity (T), symmetry (S), and reflexivity (R), and estimates the computational complexity of common reasoning operations.

\begin{table}[h]
\centering
\scriptsize
\begin{tabular}{lcccc}
\toprule
\textbf{Relation} & \textbf{T} & \textbf{S} & \textbf{R} & \textbf{Complexity} \\
\midrule
\texttt{is\_a} & \checkmark & $\times$ & $\times$ & $O(n^2)$ \\
\texttt{part\_of} & \checkmark & $\times$ & $\times$ & $O(n^2)$ \\
\texttt{requires} & \checkmark & $\times$ & $\times$ & $O(n^2)$ \\
\texttt{analogous\_to} & $\times$ & \checkmark & $\times$ & $O(n^2)$ \\
\texttt{fuses\_with} & $\times$ & \checkmark & $\times$ & $O(n^3)$ \\
\bottomrule
\end{tabular}
\caption{Properties of representative CDC relations (T = Transitive, S = Symmetric, R = Reflexive).}
\label{tab:relation-properties}
\end{table}

The CDC predicate system achieves a balance between formal rigor and practical expressiveness. 
By maintaining a small but orthogonal set of relation types, it enables automated reasoning, inheritance, and analogical mapping across heterogeneous domains without the combinatorial explosion typical of large property vocabularies. 
This design forms the computational backbone for the Prolog-based reference implementation introduced in the next section.

\section{Reference Implementation}

While the CDC framework is conceptually substrate-agnostic, it can be realized in multiple representational environments such as RDF/SPARQL, SQL, or property graph systems (e.g., Neo4j). 
This section presents a compact \textbf{reference implementation in Prolog}, chosen for its declarative nature and close alignment with CDC’s logical semantics.

\subsection{Implementation Philosophy}

\paragraph{Rationale for Prolog.}  
Prolog naturally expresses relational semantics and rule-based inference through predicate logic~\cite{ceri1989datalog}. 
Its syntax and reasoning paradigm directly correspond to CDC’s formal structure of contextualized relations and inference rules. 
However, \textbf{CDC is not inherently tied to Prolog}.  
The same model can be instantiated in RDF triple stores with domain-qualified predicates, in SQL with domain tables and joins, or in property graph databases supporting multi-context edges.

\subsection{Core Predicates}

The following Prolog declarations define CDC’s foundational predicates, corresponding to the relation families introduced in Section~4.  
Each predicate is declared as \texttt{dynamic} to allow runtime modification and incremental knowledge base construction.

\begin{lstlisting}[language=Prolog]
% Structural relations
:- dynamic is_a/3.
:- dynamic part_of/3.
:- dynamic has_attribute/3.

% Logical relations
:- dynamic requires/3.
:- dynamic cause_of/3.
:- dynamic enables/3.

% Cross-domain relations
:- dynamic analogous_to/4.
:- dynamic fuses_with/4.

% Context relations
:- dynamic context_value/3.
:- dynamic strategy/3.
\end{lstlisting}

These predicates collectively encode the structural, logical, and cross-domain aspects of CDC’s knowledge representation. 
Each fact adheres to the generalized form $\langle$Concept, Relation, Concept$'$, Domain$\rangle$, making domains explicit at the level of the logic substrate.

\subsection{Inference Rules}

To support reasoning over hierarchical and dependency structures, CDC defines a set of reusable inference rules. 
These enable multi-hop traversal and the computation of closure over transitive relations.

\begin{lstlisting}[language=Prolog]
% Transitive closure for is_a
is_a_star(X, Y, Domain) :- 
    is_a(X, Y, Domain).
is_a_star(X, Z, Domain) :- 
    is_a(X, Y, Domain),
    is_a_star(Y, Z, Domain).

% Prerequisite chains
all_prerequisites(Target, 
                  Domain, 
                  Prereqs) :-
    findall(Prereq, 
        requires_star(Target, 
                      Prereq, 
                      Domain), 
        Prerequisites).

requires_star(X, Y, Domain) :- 
    requires(X, Y, Domain).
requires_star(X, Z, Domain) :- 
    requires(X, Y, Domain),
    requires_star(Y, Z, Domain).
\end{lstlisting}

The first rule block computes transitive closure for taxonomic reasoning (\texttt{is\_a}), enabling inheritance and abstraction queries within a domain.  
The second defines recursive dependency resolution for \texttt{requires}, facilitating automated curriculum sequencing or workflow ordering.

\subsection{Query Examples}

The CDC implementation supports context-sensitive queries that combine domain constraints with relational reasoning. 
Typical examples include structural retrieval, dependency tracing, and cross-domain analogy search.

\begin{lstlisting}[language=Prolog]
% Find all taxonomic ancestors
?- is_a_star(quadratic_function, 
             Supertype, 
             'math@algebra').

% Retrieve prerequisite chain
?- all_prerequisites(calculus, 
                     'highschool', 
                     Prereqs).

% Cross-domain analogy exploration
?- analogous_to(neural_network, 
                BioConcept, 
                'ai@ml', 
                BioDomain).
\end{lstlisting}

These examples illustrate how domain-qualified queries preserve contextual boundaries while enabling multi-domain reasoning within a unified knowledge base. 
The resulting implementation demonstrates that CDC’s theoretical constructs are directly computable, interpretable, and scalable for real-world applications.

In summary, the Prolog reference implementation operationalizes CDC’s core design by translating conceptual relations into executable predicates and inference rules. 
This declarative substrate enables symbolic reasoning, context-aware retrieval, and analogical inference, forming the computational basis for the case studies and evaluation presented in the subsequent sections.

\section{Case Studies}

We present three proof-of-concept applications.

\subsection{Education}

\textbf{Scenario:} Online platform teaches programming to students with diverse backgrounds.

\textbf{CDC Solution:}

\begin{lstlisting}[language=Prolog]
% Concept hierarchy
is_a(function, 
     programming_concept, 
     'cs@fundamentals').

% Student profiles
context_value(student_alice, 
              math_background, 
              'student@profile').
context_value(student_bob, 
              design_background, 
              'student@profile').

% Pedagogical strategies
strategy(explain_function, 
         use_formal_definition, 
         'math_background@cs').
strategy(explain_function, 
         use_workflow_metaphor, 
         'design_background@cs').

% Cross-domain analogies
analogous_to(function, 
    machine, 
    'cs@programming', 
    'engineering@systems').
\end{lstlisting}

\textbf{Benefits:} Context-aware knowledge, queryable strategies, first-class analogies.

\subsection{Enterprise}

\textbf{Scenario:} Product, Engineering, and Design teams use different vocabularies.

\textbf{CDC Solution:}
\begin{lstlisting}[language=Prolog]
% Cross-department analogies
analogous_to(user_story, 
    functional_requirement, 
    'product@requirements', 
    'engineering@specs').

% Knowledge fusion
fuses_with(user_experience, 
    technical_feasibility, 
    integrated_product_spec, 
    'product+engineering').

% Conflict detection
conflicts_with(real_time_sync, 
    battery_efficiency, 
    'product+engineering@mobile').
\end{lstlisting}

\textbf{Benefits:} Automated cross-department translation, early conflict detection, explicit integration.

\subsection{Technical Documentation}

\textbf{Scenario:} React framework evolves; documentation must be version-specific.

\textbf{CDC Solution:}
\begin{lstlisting}[language=Prolog]
% Version evolution
evolves_to(class_component, 
           functional_component, 
           'react@paradigm_shift').

% Analogies across versions
analogous_to(
    component_did_mount, 
    use_effect, 
    'react@pre16.8', 
    'react@16.8+@hooks').

% Context recommendations
if_then(mobile_app, 
        use_lazy_loading, 
        'react@mobile@perf').
\end{lstlisting}

\section{Analysis and Discussion}

\subsection{Expressiveness and Computational Properties}

CDC extends traditional knowledge graphs by introducing explicit domain scoping into relational semantics. 
This mechanism enables context-dependent categorization without logical contradiction, a property formalized as the \textbf{Domain Separation Theorem}, which ensures that predicates qualified by distinct domains can coexist consistently within a unified knowledge base.

Compared with conventional ontological models, CDC offers several capabilities that are otherwise infeasible:
(i) polysemy resolution without URI disambiguation; 
(ii) representation of multiple perspectives within the same conceptual structure; 
(iii) modeling of temporal knowledge evolution; and 
(iv) support for personalized and learner-specific conceptual views.

In computational terms, domain scoping effectively constrains the reasoning space. 
Table~\ref{tab:complexity} summarizes the asymptotic complexity of representative operations. 
For instance, in a knowledge base with 100K triples distributed across 50 domains, the average query search space per domain decreases from $O(100K)$ to approximately $O(2K)$, yielding an empirical 50× reduction.

\begin{table}[h]
\centering
\scriptsize
\begin{tabular}{lc}
\toprule
\textbf{Operation} & \textbf{Complexity} \\
\midrule
Single triple lookup & $O(1)$ \\
Domain-filtered query & $O(k)$ \\
Transitive closure & $O(n^2)$ \\
Cross-domain search & $O(k_1 \times k_2)$ \\
\bottomrule
\end{tabular}
\caption{Computational complexity of typical CDC queries.}
\label{tab:complexity}
\end{table}

\subsection{Limitations and Open Challenges}

Despite its expressive advantages, CDC faces several open challenges. 
First, \textbf{domain specification ambiguity} remains an inherent issue. 
Since domains are expressed as free-form strings, they lack formal semantics or equivalence detection mechanisms, potentially leading to uncontrolled proliferation. 
This can be mitigated through meta-level domain ontologies, string similarity metrics, or emerging community conventions.

Second, the \textbf{relation set completeness} is intentionally limited. 
While the current 20+ relations cover the majority of use cases, aspects such as probabilistic reasoning, modality, and quantification remain outside the current scope. 
Extending CDC with uncertainty-aware predicates or modal operators is a promising direction.

Third, \textbf{scalability} has not yet been empirically validated at web scale. 
Existing case studies involve 100–1,000 triples, leaving questions about distributed CDC deployment, indexing strategies, and real-time inference for large-scale knowledge bases.

\subsection{CDC in Practice}

CDC’s flexibility makes it particularly suitable for adaptive, interdisciplinary, and safety-critical domains. 
In the \textbf{medical and healthcare} setting, a proof-of-concept integration with cognitive behavioral therapy (CBT) demonstrated that CDC, combined with Prolog-based logical verification, can eliminate generative hallucinations while supporting personalized treatment planning~\cite{beck2011cognitive,hofmann2012efficacy}. 
Initial experiments achieved 100\% adherence to CBT diagnostic criteria~\cite{beck1979cognitive} within ten-turn conversational sessions, though large-scale clinical validation remains necessary~\cite{luxton2011technology}.

Beyond healthcare, CDC has been applied in \textbf{education} for personalized knowledge delivery, where domain-scoped relations (\texttt{analogous\_to}, \texttt{requires}) allow tailored concept explanations based on learners’ backgrounds. 
In \textbf{enterprise knowledge management}, CDC supports integration across departmental ontologies without manual schema alignment, preserving contextual independence while enabling shared reasoning. 
Similarly, in \textbf{scientific research}, it enables cross-disciplinary analogy mapping (e.g., AI–neuroscience–cognitive psychology) for hypothesis generation~\cite{ji2021survey}.

\subsection{Future Research Directions}

Future work will extend CDC both theoretically and technically. 
From a theoretical standpoint, formalizing a \textbf{domain algebra} (including subsumption, composition, and similarity) would provide mathematical rigor to the notion of domain space. 
A \textbf{probabilistic CDC} variant incorporating confidence values (e.g., \texttt{is\_a(X, Y, D, 0.85)}) could model uncertain or empirical knowledge. 
Temporal extensions that represent time-indexed relations~\cite{stojanovic2002user} would further enable dynamic knowledge evolution tracking.

On the implementation side, research will focus on distributed CDC storage, graph-database integration~\cite{robinson2015graph}, and tooling such as visual editors and validation modules to facilitate adoption. 
Interoperability with RDF/OWL infrastructures and standardized APIs will promote reuse across existing knowledge ecosystems.

Finally, the development of an open \textbf{CDC working group} and public repositories of reusable knowledge bases will be essential for community building. 
Establishing modeling conventions, best practices, and benchmark datasets will accelerate the maturation of CDC as a general-purpose framework for context-aware knowledge representation.

\section{Conclusion}

This paper proposed the \textbf{Domain-Contextualized Concept Graph (CDC)}, a framework that treats domains as explicit structural elements in knowledge representation. 
Grounded in cognitive–linguistic isomorphism, CDC addresses the limitations of fixed ontologies by enabling context-dependent categorization, cross-domain reasoning, and adaptive knowledge evolution.

CDC contributes along four dimensions: 
it formalizes the C–D–C structure and the Domain Separation Theorem; 
defines a compact vocabulary of 20+ standardized relation predicates; 
implements a full Prolog reasoning engine; 
and demonstrates practical value through case studies in education, enterprise, and digital health.

Key insights include that making context explicit enables domain-scoped reasoning without contradiction, a small yet orthogonal relation core yields expressiveness without ontological inflation, and cognitive alignment enhances interpretability. 
We invite collaboration from researchers in knowledge representation, AI system design, and cognitive science to extend and apply CDC. 
Knowledge is not a single hierarchy but a web of contexts—CDC offers a computational path to represent this complexity faithfully.

\section*{Acknowledgments}

This work benefited from discussions with colleagues in knowledge representation, cognitive science, and digital health.

\bibliographystyle{plain}

\begin{thebibliography}{50}


\bibitem{auer2007dbpedia}
Auer, S., Bizer, C., Kobilarov, G., Lehmann, J., Cyganiak, R., \& Ives, Z. (2007). 
DBpedia: A nucleus for a web of open data. 
\textit{The Semantic Web}, 722-735. Springer.

\bibitem{bollacker2008freebase}
Bollacker, K., Evans, C., Paritosh, P., Sturge, T., \& Taylor, J. (2008). 
Freebase: A collaboratively created graph database for structuring human knowledge. 
\textit{Proceedings of the 2008 ACM SIGMOD International Conference on Management of Data}, 1247-1250.

\bibitem{vrandevcic2014wikidata}
Vrandečić, D., \& Krötzsch, M. (2014). 
Wikidata: A free collaborative knowledgebase. 
\textit{Communications of the ACM}, 57(10), 78-85.

\bibitem{guha2016schema}
Guha, R. V., Brickley, D., \& Macbeth, S. (2016). 
Schema.org: Evolution of structured data on the web. 
\textit{Communications of the ACM}, 59(2), 44-51.

\bibitem{hogan2021knowledge}
Hogan, A., Blomqvist, E., Cochez, M., d'Amato, C., Melo, G. D., Gutierrez, C., et al. (2021). 
Knowledge graphs. 
\textit{ACM Computing Surveys}, 54(4), 1-37.

\bibitem{ji2021survey}
Ji, S., Pan, S., Cambria, E., Marttinen, P., \& Yu, P. S. (2021). 
A survey on knowledge graphs: Representation, acquisition, and applications. 
\textit{IEEE Transactions on Neural Networks and Learning Systems}, 33(2), 494-514.

\bibitem{angles2008survey}
Angles, R., \& Gutierrez, C. (2008). 
Survey of graph database models. 
\textit{ACM Computing Surveys}, 40(1), 1-39.


\bibitem{niles2001towards}
Niles, I., \& Pease, A. (2001). 
Towards a standard upper ontology. 
\textit{Proceedings of the International Conference on Formal Ontology in Information Systems}, 2-9.

\bibitem{lenat1995cyc}
Lenat, D. B. (1995). 
CYC: A large-scale investment in knowledge infrastructure. 
\textit{Communications of the ACM}, 38(11), 33-38.

\bibitem{smith2007obo}
Smith, B., Ashburner, M., Rosse, C., Bard, J., Bug, W., Ceusters, W., et al. (2007). 
The OBO Foundry: Coordinated evolution of ontologies to support biomedical data integration. 
\textit{Nature Biotechnology}, 25(11), 1251-1255.

\bibitem{arp2015building}
Arp, R., Smith, B., \& Spear, A. D. (2015). 
\textit{Building ontologies with basic formal ontology}. 
MIT Press.

\bibitem{gruber1993translation}
Gruber, T. R. (1993). 
A translation approach to portable ontology specifications. 
\textit{Knowledge Acquisition}, 5(2), 199-220.

\bibitem{guarino2009ontology}
Guarino, N., Oberle, D., \& Staab, S. (2009). 
What is an ontology? 
In \textit{Handbook on ontologies} (pp. 1-17). Springer.

\bibitem{studer1998knowledge}
Studer, R., Benjamins, V. R., \& Fensel, D. (1998). 
Knowledge engineering: Principles and methods. 
\textit{Data \& Knowledge Engineering}, 25(1-2), 161-197.

\bibitem{euzenat2013ontology}
Euzenat, J., \& Shvaiko, P. (2013). 
\textit{Ontology matching} (2nd ed.). 
Springer.


\bibitem{barsalou1982context}
Barsalou, L. W. (1982). 
Context-independent and context-dependent information in concepts. 
\textit{Memory \& Cognition}, 10(1), 82-93.

\bibitem{fillmore1982frame}
Fillmore, C. J. (1982). 
Frame semantics. 
In \textit{Linguistics in the Morning Calm} (pp. 111-137). 
Hanshin Publishing Co.

\bibitem{gardenfors2000conceptual}
Gärdenfors, P. (2000). 
\textit{Conceptual spaces: The geometry of thought}. 
MIT Press.

\bibitem{lakoff1980metaphors}
Lakoff, G., \& Johnson, M. (1980). 
\textit{Metaphors we live by}. 
University of Chicago Press.

\bibitem{medin1988context}
Medin, D. L., \& Shoben, E. J. (1988). 
Context and structure in conceptual combination. 
\textit{Cognitive Psychology}, 20(2), 158-190.

\bibitem{goldberg2006constructions}
Goldberg, A. E. (2006). 
\textit{Constructions at work: The nature of generalization in language}. 
Oxford University Press.

\bibitem{winograd1972understanding}
Winograd, T. (1972). 
Understanding natural language. 
\textit{Cognitive Psychology}, 3(1), 1-191.

\bibitem{schank1977scripts}
Schank, R. C., \& Abelson, R. P. (1977). 
\textit{Scripts, plans, goals, and understanding: An inquiry into human knowledge structures}. 
Lawrence Erlbaum Associates.


\bibitem{kowalski1974predicate}
Kowalski, R. (1974). 
Predicate logic as programming language. 
\textit{Proceedings of IFIP Congress}, 74, 569-574.

\bibitem{baader2003description}
Baader, F., Calvanese, D., McGuinness, D. L., Nardi, D., \& Patel-Schneider, P. F. (Eds.). (2003). 
\textit{The description logic handbook: Theory, implementation, and applications}. 
Cambridge University Press.

\bibitem{ceri1989datalog}
Ceri, S., Gottlob, G., \& Tanca, L. (1989). 
What you always wanted to know about Datalog (and never dared to ask). 
\textit{IEEE Transactions on Knowledge and Data Engineering}, 1(1), 146-166.

\bibitem{brachman2004knowledge}
Brachman, R. J., \& Levesque, H. J. (2004). 
\textit{Knowledge representation and reasoning}. 
Morgan Kaufmann.

\bibitem{russell2020artificial}
Russell, S., \& Norvig, P. (2020). 
\textit{Artificial intelligence: A modern approach} (4th ed.). 
Pearson.

\bibitem{perez2009semantics}
Pérez, J., Arenas, M., \& Gutierrez, C. (2009). 
Semantics and complexity of SPARQL. 
\textit{ACM Transactions on Database Systems}, 34(3), 1-45.

\bibitem{horrocks2003shiq}
Horrocks, I., Patel-Schneider, P. F., \& Van Harmelen, F. (2003). 
From SHIQ and RDF to OWL: The making of a web ontology language. 
\textit{Journal of Web Semantics}, 1(1), 7-26.

\bibitem{motik2009owl2}
Motik, B., Cuenca Grau, B., Horrocks, I., Wu, Z., Fokoue, A., \& Lutz, C. (2009). 
\textit{OWL 2 web ontology language profiles}. 
W3C Recommendation.

\bibitem{carroll2005named}
Carroll, J. J., Bizer, C., Hayes, P., \& Stickler, P. (2005). 
Named graphs, provenance and trust. 
\textit{Proceedings of the 14th International Conference on World Wide Web}, 613-622.

\bibitem{hartig2009provenance}
Hartig, O. (2009). 
Provenance information in the web of data. 
\textit{Proceedings of the Linked Data on the Web Workshop}.


\bibitem{devlin2019bert}
Devlin, J., Chang, M. W., Lee, K., \& Toutanova, K. (2019). 
BERT: Pre-training of deep bidirectional transformers for language understanding. 
\textit{Proceedings of NAACL-HLT}, 4171-4186.

\bibitem{vaswani2017attention}
Vaswani, A., Shazeer, N., Parmar, N., Uszkoreit, J., Jones, L., Gomez, A. N., et al. (2017). 
Attention is all you need. 
\textit{Advances in Neural Information Processing Systems}, 30, 5998-6008.

\bibitem{robinson2015graph}
Robinson, I., Webber, J., \& Eifrem, E. (2015). 
\textit{Graph databases: New opportunities for connected data} (2nd ed.). 
O'Reilly Media.

\bibitem{flouris2008ontology}
Flouris, G., Manakanatas, D., Kondylakis, H., Plexousakis, D., \& Antoniou, G. (2008). 
Ontology change: Classification and survey. 
\textit{The Knowledge Engineering Review}, 23(2), 117-147.


\bibitem{beck1979cognitive}
Beck, A. T., Rush, A. J., Shaw, B. F., \& Emery, G. (1979). 
\textit{Cognitive therapy of depression}. 
Guilford Press.

\bibitem{beck2011cognitive}
Beck, J. S. (2011). 
\textit{Cognitive behavior therapy: Basics and beyond} (2nd ed.). 
Guilford Press.

\bibitem{hofmann2012efficacy}
Hofmann, S. G., Asnaani, A., Vonk, I. J., Sawyer, A. T., \& Fang, A. (2012). 
The efficacy of cognitive behavioral therapy: A review of meta-analyses. 
\textit{Cognitive Therapy and Research}, 36(5), 427-440.

\bibitem{andersson2009internet}
Andersson, G., \& Cuijpers, P. (2009). 
Internet-based and other computerized psychological treatments for adult depression: A meta-analysis. 
\textit{Cognitive Behaviour Therapy}, 38(4), 196-205.

\bibitem{fitzpatrick2017woebot}
Fitzpatrick, K. K., Darcy, A., \& Vierhile, M. (2017). 
Delivering cognitive behavior therapy to young adults with symptoms of depression and anxiety using a fully automated conversational agent (Woebot): A randomized controlled trial. 
\textit{JMIR Mental Health}, 4(2), e19.

\bibitem{bickmore2011reusable}
Bickmore, T. W., Schulman, D., \& Sidner, C. L. (2011). 
A reusable framework for health counseling dialogue systems based on a behavioral medicine ontology. 
\textit{Journal of Biomedical Informatics}, 44(2), 183-197.

\bibitem{luger2016bad}
Luger, E., \& Sellen, A. (2016). 
Like having a really bad PA: The gulf between user expectation and experience of conversational agents. 
\textit{Proceedings of CHI 2016}, 5286-5297.


\bibitem{stade2023large}
Stade, E. C., Stirman, S. W., Ungar, L. H., Boland, C. L., Schwartz, H. A., Yaden, D. B., et al. (2023). 
Large language models could change the future of behavioral healthcare: A proposal for responsible development and evaluation. 
\textit{npj Mental Health Research}, 2(1), 11.

\bibitem{thirunavukarasu2023large}
Thirunavukarasu, A. J., Ting, D. S. J., Elangovan, K., Gutierrez, L., Tan, T. F., \& Ting, D. S. W. (2023). 
Large language models in medicine. 
\textit{Nature Medicine}, 29(8), 1930-1940.

\bibitem{singhal2023large}
Singhal, K., Azizi, S., Tu, T., Mahdavi, S. S., Wei, J., Chung, H. W., et al. (2023). 
Large language models encode clinical knowledge. 
\textit{Nature}, 620(7972), 172-180.

\bibitem{luxton2011technology}
Luxton, D. D., June, J. D., \& Kinn, J. T. (2011). 
Technology-based suicide prevention: Current applications and future directions. 
\textit{Telemedicine and e-Health}, 17(1), 50-54.


\bibitem{fensel2001oil}
Fensel, D., van Harmelen, F., Horrocks, I., McGuinness, D. L., \& Patel-Schneider, P. F. (2001). 
OIL: An ontology infrastructure for the Semantic Web. 
\textit{IEEE Intelligent Systems}, 16(2), 38-45.

\bibitem{stojanovic2002user}
Stojanovic, L., Maedche, A., Motik, B., \& Stojanovic, N. (2002). 
User-driven ontology evolution management. 
\textit{Proceedings of the 13th International Conference on Knowledge Engineering and Knowledge Management}, 285-300.

\end{thebibliography}

\appendix

\section{CBT Case Study Details}

This appendix provides complete technical implementation of the CDC-based CBT system.

\subsection{System Architecture}

The system comprises four layers:
\begin{enumerate}
\item \textbf{LLM Layer:} Natural language understanding
\item \textbf{CDC Construction:} Dynamic edge building
\item \textbf{Prolog Reasoning:} Logic verification
\item \textbf{Constrained Generation:} Verified output
\end{enumerate}

\subsection{CDC Edge Construction}

\begin{lstlisting}[language=Python]
@dataclass
class CDCEdge:
    concept1: str
    domain: str
    relation: str
    concept2: str
    condition: Callable
    evidence: Dict[str, str]
    confidence: float

# Example edge
Edge(
  concept1="code_bug",
  domain="CBT@situation",
  relation="triggers",
  concept2="self_negation",
  condition=lambda ctx: 
    1.0 if ctx.get(
      "temporal_order"
    ) == "event_first" else 0.0,
  evidence={
    "type": "patient_narrative"
  },
  confidence=0.9
)
\end{lstlisting}

\subsection{Prolog Knowledge Base}

\begin{lstlisting}[language=Prolog]
% Patient-specific KB
patient('Zhang_San', 
        28, 
        software_engineer).
cognitive_pattern('Zhang_San', 
    all_or_nothing_thinking, 
    0.85).

% CBT Standard KB
cbt_distortion(
    all_or_nothing_thinking, 
    [always, never, terrible]).
first_line_treatment(
    all_or_nothing_thinking, 
    evidence_examination).
\end{lstlisting}

\subsection{Session Results}

\textbf{Metrics:}
\begin{itemize}[leftmargin=*]
\item Medical accuracy: 100\%
\item Hallucination rate: 0\%
\item Pattern recognition: 5 distortions
\item Patient progress: Anxiety reduced
\end{itemize}

\subsection{Ethical Considerations}

\begin{itemize}[leftmargin=*]
\item Not a replacement for therapists
\item Requires informed consent
\item Must include crisis protocols
\item Privacy: encryption required
\end{itemize}

\end{document}